\documentclass{article}
\usepackage{spconf,amsmath,graphicx,hyperref}
\usepackage{xcolor}
\usepackage{placeins}
\usepackage{tabularx,booktabs}
\usepackage{caption}
\renewcommand{\paragraph}[1]{\noindent\textbf{#1}\quad}

\title{PersonaPlex: Voice and Role Control for Full Duplex Conversational Speech Models}
\name{
  \begin{tabular}{c}
    Rajarshi Roy, Jonathan Raiman, Sang-gil Lee, Teodor-Dumitru Ene, \\
    Robert Kirby, Sungwon Kim, Jaehyeon Kim, Bryan Catanzaro
  \end{tabular}
}
\address{
  NVIDIA \\
  {\small \texttt{\{rajarshir, jraiman, sanggill, tene, rkirby, sungwonk, jaehyeonk, bcatanzaro\}@nvidia.com}}
}

\usepackage{fancyhdr}

\begin{document}
%
\maketitle

\thispagestyle{fancy}
\fancyhf{}
\fancyfoot[C]{{\small\textit{Preprint. Under review at ICASSP 2026.}}}
\renewcommand{\headrulewidth}{0pt}
\renewcommand{\footrulewidth}{0pt}

\begin{abstract}
Recent advances in duplex speech models have enabled natural, low-latency speech-to-speech interactions. However, existing models are restricted to a fixed role and voice, limiting their ability to support structured, role-driven real-world applications and personalized interactions. In this work, we introduce PersonaPlex, a duplex conversational speech model that incorporates hybrid system prompts, combining role conditioning with text prompts and voice cloning with speech samples. PersonaPlex is trained on a large-scale synthetic dataset of paired prompts and user-agent conversations, generated with open-source large language models (LLM) and text-to-speech (TTS) models. To evaluate role conditioning in real-world settings, we extend the {\it Full-Duplex-Bench} benchmark beyond a single assistant role to multi-role customer service scenarios. Experiments show that PersonaPlex achieves strong role-conditioned behavior, voice-conditioned speech, and natural conversational responsiveness, surpassing state-of-the-art duplex speech models and hybrid large language model–based speech systems in role adherence, speaker similarity, latency, and naturalness.

\end{abstract}
%


\section{Introduction}
\label{sec:intro}
Recent advances in duplex speech models have enabled real-time, low-latency speech-to-speech conversation systems that approximate natural human interaction. These systems integrate ASR, LLMs, and TTS into a unified pipeline, enabling agents to respond with high naturalness and low turn-taking delay. However, current duplex systems are generally limited to a fixed voice identity and role, restricting their use in structured or role-driven applications such as customer service, multi-character interactions, and personalized assistants. Without the ability to condition on conversational role and speaker identity, these systems fail to achieve the flexibility required in real-world human–machine interactions.

In parallel, research on voice-conditioned TTS has demonstrated progress in speaker adaptation, voice cloning, and prosody manipulation, while instruction-following LLMs have shown strong role conditioning in text-based interactions. However, these advances have not been fully realized in duplex speech systems, where latency constraints and coupled speech–text dynamics make conditioning on both role and voice more challenging. Bridging this gap requires a speech-to-speech model that integrates the conditioning capabilities of LLMs and the adaptability of modern TTS into a low-latency duplex framework.

In this work, we present PersonaPlex, a full duplex speech-to-speech conversational model that incorporates hybrid system prompts, combining text-based role conditioning with audio-based voice cloning. PersonaPlex enables zero-shot voice cloning and fine-grained role conditioning, extending duplex speech beyond generic assistants to structured domains such as customer service. We introduce a large-scale synthetic training corpus of paired prompts and user-agent conversations. We evaluate PersonaPlex on the {\it Full-Duplex-Bench} benchmark \cite{lin2025fullduplexbenchbenchmarkevaluatefullduplex}. Since {\it Full-Duplex-Bench} is limited to a single assistant role, we propose {\it Service-Duplex-Bench}, an extension that covers real-world multi-role customer service scenarios. Our extension adds 350 customer service evaluation questions - each corresponding to a specific service role - to the 400 questions in {\it Full-Duplex-Bench}. In our experiments on {\it Full-Duplex-Bench} and {\it Service-Duplex-Bench} we find that PersonaPlex achieves state-of-the-art performance in role adherence, voice similarity, and dialog naturalness, while maintaining the responsiveness and turn taking abilities of duplex speech models.
\begin{figure*}[!t]
\centering
\includegraphics[width=0.9\linewidth]{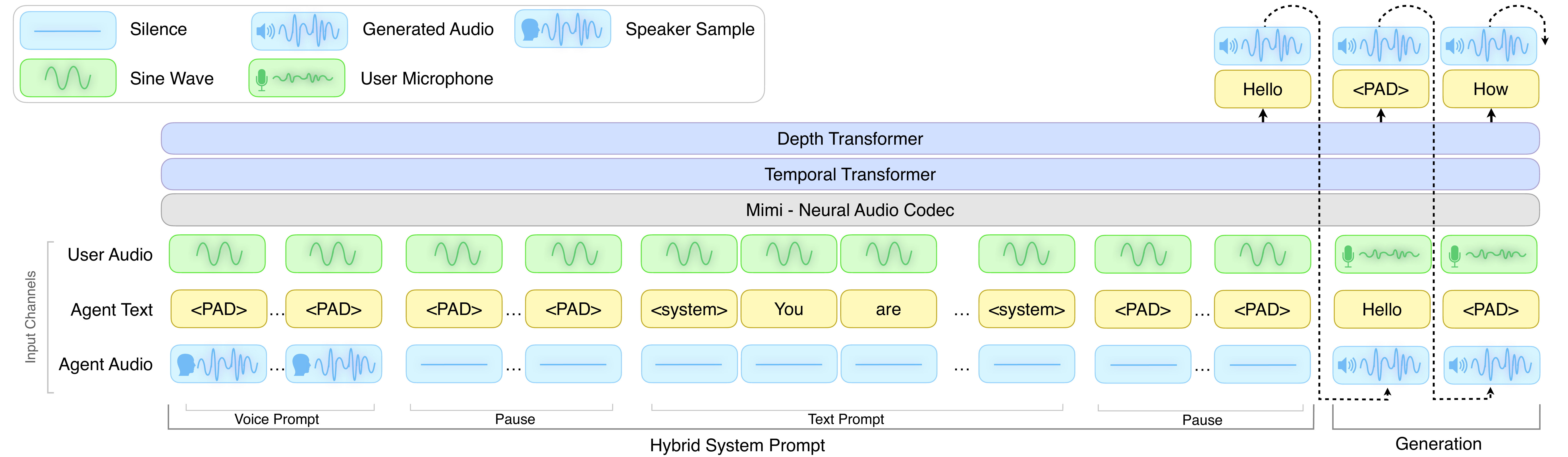}
\caption{PersonaPlex's neural network is a duplex speech model based on Moshi~\cite{moshi} with a Hybrid System Prompt enabling textual prompts and voice cloning. The model then autoregressively generates text and audio while receiving live user audio.}
\label{fig:architecture}
\end{figure*}

\section{Related Work}
\label{sec:related}
Cascaded ASR-LLM-TTS systems leverage the reasoning ability of LLMs and the naturalness of dedicated TTS models, but inevitably lose paralinguistic information, reducing dialog naturalness. To address this, several approaches have been proposed to improve conversational quality. Streaming TTS models~\cite{minicpm, kyutai_tts} reduce LLM$\rightarrow$TTS latency and support voice cloning. HumeAI's Emphatic Voice Interface\footnote{\url{https://www.hume.ai/empathic-voice-interface}} encodes paralinguistic information in text to preserve expressive knowledge. Half-duplex approaches~\cite{mini_omni, mini_omni2, qwen2.5-omni, llama-omni2, glm-4-voice} directly consume and stream speech tokens to lower latency and retain paralinguistics. However, these models remain reliant on external turn-taking mechanisms, do not listen while speaking, and are limited to fixed voices.  

Full-duplex models~\cite{moshi, omni_flatten, syncllm, salm_duplex} support real-time speech-to-speech interaction with natural turn taking and responsiveness, but remain limited to fixed voices and assistant-style roles. Commercial systems~\cite{gpt_realtime, gemini_live} allow role conditioning via context prompts, yet voices are still fixed. Our work integrates duplex modeling, zero-shot voice cloning, and instruction-based role conditioning to address these limitations and broaden applicability to real-world settings.

\paragraph{Duplex Speech Model Evaluation} Most conversational speech evaluation benchmarks~\cite{chen2024voicebench,cheng2025voxdialogue,cui2025voxeval,yan2025uro} focus on single-turn responses and enforce a half-duplex turn-by-turn interaction. The {\it Full-Duplex-Bench} \cite{lin2025fullduplexbenchbenchmarkevaluatefullduplex} benchmark follows a full-duplex setup where user input audio is streamed in and the model's corresponding generated audio is captured and evaluated. This benchmark evaluates both knowledge and reasoning capabilities with QA tasks and testing conversational dynamics capabilities such as turn-taking, responsiveness, interruption and backchannel handling.
To better evaluate fine-grained role conditioning in real-world applications, we propose an extension to {\it Full-Duplex-Bench} that covers customer service role evaluations.


\section{PersonaPlex}
\label{sec:personaplex}

\subsection{Architecture}
\label{ssec:architecture}
We propose PersonaPlex, a duplex-style multimodal model that follows the Moshi~\cite{moshi} architecture by receiving three input streams: {\it user audio}, {\it agent text}, and {\it agent audio}. An overview of the architecture is shown in \autoref{fig:architecture}. To jointly enable role conditioning and voice control, we introduce a {\it Hybrid System Prompt} combining textual role descriptions with audio voice examples.

The {\it Hybrid System Prompt} consists of two temporally concatenated segments: a text prompt segment and a voice prompt segment. The text prompt segment performs role conditioning by forcing scenario-specific text tokens on the agent text channel while keeping the agent audio channel silent. The voice prompt segment performs voice prompting by supplying a short speech sample on the agent audio channel while padding the agent text channel. With this setup, subsequent agent utterances are generated in the same voice, enabling zero-shot voice cloning. For stable conditioning, the user audio channel is replaced with a 440 Hz sine wave, and custom text/audio delimiters mark the boundary between the Hybrid System Prompt and dialogue.

We observe no difference in model performance regardless of whether the voice prompt segment or text prompt segment is positioned first. In our implementation, the voice prompt precedes the text prompt to enable prefilling during inference when zero-shot voice cloning is not required, thereby reducing latency.

During training, we mask out loss backpropagation to the system prompt. Following Moshi~\cite{moshi}, we also adjust the training objective to account for token imbalance. We down-weight the loss on non-semantic audio tokens by 0.02 and on padded text tokens by 0.3.

\begin{table*}[!t]
\centering
\caption{Dialog Naturalness MOS (95\% C.I.) and Voice Cloning Speaker Similarity.}
\label{tab:amt_naturalness}
\begin{tabular}{lccc}
\toprule
\textbf{Model} &
\shortstack{DMOS($\uparrow$)\\(Full-Duplex-Bench) } &
\shortstack{DMOS($\uparrow$)\\(Service-Duplex-Bench)} &
\shortstack{SSIM($\uparrow$)\\(Full-Duplex-Bench)} \\
\hline
PersonaPlex   & $\textbf{3.90} \pm\textbf{0.15}$ & $\textbf{3.59} \pm \textbf{0.12}$ & {\bf 0.57} \\
Gemini~\cite{gemini_live}     & $3.72 \pm 0.14$ & $3.22 \pm 0.14$ & 0.00 \\
Qwen-2.5-Omni~\cite{qwen2.5-omni} & $3.70 \pm 0.13$ & $2.37 \pm 0.20$ & 0.07 \\
Freeze-Omni~\cite{wang2024freeze}   & $3.51 \pm 0.18$ & $2.38 \pm 0.21$ & 0.05 \\
Moshi~\cite{moshi}         & $3.11 \pm 0.15$ & $2.83 \pm 0.13$ & 0.10 \\
\bottomrule
\end{tabular}
\end{table*}

\begin{table*}[!t]
\centering
\caption{Full Duplex Bench Benchmark Results}
\label{tab:full_duplex_bench}
\resizebox{\textwidth}{!}{
\begin{tabular}{lcccccccccc}
\toprule
& \multicolumn{1}{c}{\textbf{Pause (Synthetic)}} & \multicolumn{1}{c}{\textbf{Pause (Candor)}} & \multicolumn{3}{c}{\textbf{Backchannel}} & \multicolumn{2}{c}{\textbf{Smooth Turn Taking}} & \multicolumn{3}{c}{\textbf{User Interruption}} \\
\cmidrule(lr){2-3} \cmidrule(lr){4-6} \cmidrule(lr){7-8} \cmidrule(lr){9-11}
 \textbf{Model} & TOR (↓) & TOR (↓) & TOR (↓) & Freq (↑) & JSD (↓) & TOR (↑) & Latency (↓) & TOR (↑) & GPT-4o (↑) & Latency (↓) \\
\midrule
PersonaPlex  & 0.584 & 0.662 & 0.327 & \textbf{0.025} & \textbf{0.649} & \textbf{0.992} & \textbf{0.070} & \textbf{1.000} & 4.210 & 0.400 \\
Qwen-2.5-Omni   & -   & -       &  -     &  -     &  -     &  -     &  -  & -    & \textbf{4.590} & 2.740    \\
Freeze-Omni  & 0.642 & 0.481 & 0.636 & 0.001 & 0.997 & 0.336 & 0.953 & 0.867 & 3.615 & 1.409 \\
Gemini       & \textbf{0.255} & \textbf{0.310} & \textbf{0.091} & 0.012 & 0.896 & 0.655 & 1.301 & 0.891 & 3.376 & 1.183 \\
Moshi        & 0.985 & 0.980 & 1.000 & 0.001 & 0.957 & 0.941 & 0.265 & 1.000 & 0.765 & \textbf{0.257} \\
dGSLM        & 0.934 & 0.935 & 0.691 & 0.015 & 0.934 & 0.975 & 0.352 & 0.917 & 0.201 & 2.531 \\
\bottomrule
\end{tabular}
}
\end{table*}

\subsection{Synthetic Data}
\label{ssec:syntheticdata}

\subsubsection{Dialog Transcripts and Text Prompt Generation}
\label{ssec:transcripts}

We construct a diverse set of synthetic dialogs reflecting the breadth of interactions encountered in two-speaker conversations. All dialog transcripts are generated using Qwen-3-32B~\cite{qwen3technicalreport} and GPT-OSS-120B~\cite{openai2025gptoss120bgptoss20bmodel}.

\paragraph{Service Scenarios} 
Dialogs are generated hierarchically by first sampling a service domain (e.g. restaurant, bank), then selecting a scenario (e.g., refund, information request, general enquiry). Each scenario is grounded with a high-level description, which is subsequently expanded into a full two-speaker transcript through large language model generation. A corresponding role context (example in \autoref{tab:service_task_small}) is generated for the service agent. Note that all training scenarios are distinct from those used in our {\it Service-Duplex-Bench} evaluation (Section~\ref{sec:serviceduplexbench}), ensuring the model is tested on unseen service contexts.

\paragraph{Question-Answering Assistant Scenarios} 
We additionally synthesize two-turn question–answering dialogs across various topics and second-question scenarios (topic change, follow up etc.). We use a fixed role for this dataset: ``You are a wise and friendly teacher. Answer questions or provide advice in a clear and engaging way.''.

\subsubsection{Dialog Speech and Voice Prompt Generation}
\label{ssec:speechgen}

\paragraph{Voice Samples} We use 26,296 single-speaker voice samples from VoxCeleb~\cite{nagrani2017voxceleb}, Libriheavy~\cite{kang2024libriheavy}, LibriTTS~\cite{zen2019libritts}, CommonAccent~\cite{demirsahin2020open}, and Fisher~\cite{cieri2004fisher} to generate synthetic dialog speech and corresponding agent voice prompts. A test set of 2,630 voice samples are reserved for speaker similarity measurements as reported in Section~\ref{ssec:ssimnaturalness}.

\paragraph{Service Scenarios} To generate natural-sounding dialog audio, we use a multispeaker TTS model that jointly generates speech for both speakers, better capturing timing, interruptions, and room tone. We selected Dia~\cite{diatts2025} as it can receive two speaker samples and generate audio continuation following a transcript while cloning each voice.

\paragraph{Question-Answering Scenarios}
Audio for question answering scenarios is generated by providing each round of dialog to Chatterbox TTS~\cite{chatterboxtts2025}. As this model supports zero-shot voice cloning, we provide a randomly chosen voice sample for each role in the generated dialog. Because Chatterbox TTS is a single-speaker TTS model, an additional audio stitching step is required. When combining the ``user" and ``agent" dialogue turns, we can choose to add additional silence padding to simulate natural turn-taking. We observe that inserting negative-duration silence instead simulates barge-in and interruption.
Prior work validates our methodology~\cite{salm_duplex}.

\begin{table}[!b]
\centering
\caption{Service-Duplex-Bench Example}
\label{tab:service_task_small}
\begingroup
\setlength{\tabcolsep}{0.1em}
\renewcommand{\arraystretch}{1.1}
\footnotesize
\begin{minipage}{0.95\columnwidth}
\textbf{Context:} You are an agent named \textbf{Brody Murphy} working for 
\textbf{National Health Coverage}, a health insurance provider. 
The customer's SSN to verify is \textbf{076-65-0542}. 
Available plans include: Basic (\$200/month), Premium (\$450/month), 
and Family (\$700/month). Enrollment requires 48 hours. 
\end{minipage}
\par\smallskip

\begin{tabularx}{\columnwidth}{@{}l l X@{}}
\toprule
\textbf{ID} & \textbf{Tag} & \textbf{User Utterance} \\
\midrule
Q0 & Proper Noun & ``Hi, could you tell me which insurance provider I'm speaking with?'' \\
Q1 & Context details & ``Can you confirm whether my Social Security Number on file is 076-75-0542?'' \\
Q2 & Context details & ``I'm interested in a plan that covers dental and vision; which of your available plans would include those benefits?'' \\
Q3 & Unfulfillable Request & ``Can you immediately enroll me in the Premium Plan and have my coverage start this afternoon?'' \\
Q4 & Customer Rudeness & ``What's the point of having a health insurance plan anyway? This whole thing is a waste of time.'' \\
Q5 & Unspecified & ``Do you have any information about the eligibility criteria for Medicare supplement plans?'' \\
Q6 & Unrelated & ``Do you offer any services for repairing household appliances or home cleaning?'' \\
\bottomrule
\end{tabularx}
\endgroup
\end{table}

\subsection{Service-Duplex-Bench}
\label{sec:serviceduplexbench}
Our extension to {\it Full-Duplex-Bench} consists of 50 unique service role scenarios with 7 questions in each scenario. Unlike the multi-turn conversational training data, each evaluation question is a single-turn probe designed to test specific capabilities such as proper noun recall, context adherence, unfulfillable request handling, and customer rudeness management within a given service context (see \autoref{tab:service_task_small} for an example). We plan to release this dataset to provide an evaluation framework for future models.

\section{Experiments and results}
\label{sec:experiments}
We train PersonaPlex by first initializing neural network weights to those of Moshi~\cite{moshi}, followed by fine-tuning using our hybrid system prompt on synthetic dialogs generated using the approach presented in Section~\ref{ssec:syntheticdata}. The full training dataset has 1840 hours of customer service dialog interactions across 105,410 dialogs, and 410 hours of general Question-Answering dialogs across 39,322 dialogs.

We train using Adam~\cite{kingma2017adammethodstochasticoptimization} with cosine annealing. The depth transformer's learning rate is 4e-6 and the temporal transformer is 2e-6. We train for 24,576 steps using a batch size of 32 with a maximum sequence length of 2048 tokens which corresponds to 163.84 seconds. Training takes 6 hours on 8xA100 GPUs.

\subsection{Dialog Naturalness and Voice Cloning}
\label{ssec:ssimnaturalness}
A primary goal of this work is achieving dialog generation that has a natural voice and fluid conversational flow. We measure naturalness by collecting a Dialogue Mean Opinion Score (DMOS) with human evaluators selected on Amazon Mechanical Turk (AMT)~\cite{ribeiro2011crowdmos}.
Evaluators rate 8 audio samples, randomized from a selection of 5 models as seen in ~\autoref{tab:amt_naturalness}, on a scale of 1 to 5.
For {\it Service-Duplex-Bench}, we poll 202 evaluators for a total of 1616 samples, while for the {\it Full-Duplex-Bench} ``User Interruption'' category we poll 152 evaluators for a total of 1216 samples. 

We also evaluate speaker similarity on Full-Duplex-Bench using the WavLM-TDNN~\cite{wavlm} speaker verification model. Specifically, we measure the cosine similarity between embeddings of each provided voice prompt and the synthesized agent speech. As shown in \autoref{tab:amt_naturalness}, PersonaPlex achieves consistently higher similarity than other baseline models, demonstrating effective voice control.

\subsection{Full-Duplex-Bench \& Service-Duplex-Bench}
We benchmark PersonaPlex against other state-of-the-art models on
{\it Full-Duplex-Bench}~\cite{lin2025fullduplexbenchbenchmarkevaluatefullduplex} in \autoref{tab:full_duplex_bench}
and {\it Service-Duplex-Bench} in \autoref{tab:service_task}
\footnote{Our evaluations of Qwen-2.5-Omni use Freeze-Omni's Voice Activity Detector (VAD) as none was originally provided.}.
PersonaPlex shows state-of-the-art performance on metrics related to human-like user interactivity.
Furthermore, on the {\it Service-Duplex-Bench} benchmark, which focuses on role adherence and instruction following, PersonaPlex outperforms all models except Gemini Live.

Combined with the human-rated naturalness evaluation, shown in \autoref{tab:amt_naturalness}, this suggests PersonaPlex has both the human-like interactivity of a full duplex model as well as the instruction-following ability of non-duplex model architectures.

\begin{table}
\centering
\caption{Service-Duplex-Bench Results}
\label{tab:service_task}
\addtolength{\tabcolsep}{-0.25em}
\begin{tabular}{lccccccc|c}
\toprule
 & \multicolumn{8}{c}{\textbf{Task GPT-4o}~$\uparrow$} \\
\cmidrule(lr){2-9}
\textbf{Model} & \textbf{Q0} & \textbf{Q1} & \textbf{Q2} & \textbf{Q3} & \textbf{Q4} & \textbf{Q5} & \textbf{Q6} & \textbf{Mean} \\
\midrule
Gemini          & 4.6 & 4.7 & 4.8 & 4.9 & 4.5 & 4.7 & 4.9 & 4.73 \\
PersonaPlex     & 4.6 & 4.6 & 4.4 & 4.5 & 4.5 & 4.3 & 4.5 & 4.48 \\
Freeze-Omni     & 3.9 & 3.5 & 3.8 & 4.3 & 4.1 & 4.2 & 4.3 & 4.02 \\
Qwen-2.5-Omni   & 1.3 & 1.6 & 2.6 & 3.4 & 3.3 & 3.6 & 3.5 & 2.76 \\
Moshi           & 1.5 & 1.4 & 1.8 & 2.0 & 1.9 & 2.1 & 1.6 & 1.75 \\
\bottomrule
\end{tabular}
\end{table}
\begin{table}
  \centering
  \caption{Dataset size effect on PersonaPlex performance.}
  \label{tab:datasetscale}
\addtolength{\tabcolsep}{-0.27em}
  \begin{tabular}{@{}rccc@{}}
    \toprule
    & \multicolumn{2}{c}{Full-Duplex-Bench} & Service-Duplex-Bench\\
\cmidrule(lr){2-3} \cmidrule(lr){4-4}
    Dataset Size & 
SSIM ($\uparrow$) & GPT-4o ($\uparrow$) & GPT-4o ($\uparrow$) \\
    \midrule
    100\% & {\bf0.57} & 4.21 & {\bf 4.48}\\
    50\% & 0.56 & {\bf 4.52} & 4.24\\
    25\% & 0.54 & 4.44 & 4.20\\
     (Moshi) 0\% & 0.10 & 0.77 & 1.75\\
    \bottomrule
  \end{tabular}
\end{table}
\subsection{Dataset Scale}
We measure the effect of dataset size by training with varying amounts of data. Adding synthetic data greatly enhances both voice cloning and role adherence versus the Moshi baseline. On {\it Full-Duplex-Bench}, strong performance is achieved with limited data, while on {\it Service-Duplex-Bench}, role adherence improves steadily with more data.

\begin{table*}[!ht]
\centering
\caption{Released Checkpoint: Full Duplex Bench Results}
\label{tab:full_duplex_bench_release}
\resizebox{\textwidth}{!}{
\begin{tabular}{lcccccccccc}
\toprule
& \multicolumn{1}{c}{\textbf{Pause (Synthetic)}} & \multicolumn{1}{c}{\textbf{Pause (Candor)}} & \multicolumn{3}{c}{\textbf{Backchannel}} & \multicolumn{2}{c}{\textbf{Smooth Turn Taking}} & \multicolumn{3}{c}{\textbf{User Interruption}} \\
\cmidrule(lr){2-3} \cmidrule(lr){4-6} \cmidrule(lr){7-8} \cmidrule(lr){9-11}
\textbf{Model} & TOR (↓) & TOR (↓) & TOR (↓) & Freq (↑) & JSD (↓) & TOR (↑) & Latency (↓) & TOR (↑) & GPT-4o (↑) & Latency (↓) \\
\midrule
PersonaPlex (Released) & 0.358 & 0.431 & 0.273 & 0.042 & 0.662 & 0.908 & 0.170 & 0.950 & 4.290 & 0.240 \\
\bottomrule
\end{tabular}
}
\end{table*}

\section{Conclusion}
\label{sec:Conclusion}

In this work, we presented PersonaPlex, a full duplex speech-to-speech conversational model that enables zero-shot voice cloning and fine-grained role conditioning through hybrid text–audio system prompts. Our results demonstrate that conditioning can be integrated into duplex speech systems without altering their underlying architecture. PersonaPlex outperforms prior duplex baselines in speaker similarity, role adherence, and dialog naturalness. To our knowledge it is the first open model to reach comparable naturalness as closed commercial systems. Our findings suggest hybrid prompt conditioning offers a scalable path toward personalized, role-conditioned conversational agents. Future work will explore post-training alignment and integration with external tools. We believe PersonaPlex advances duplex speech towards real-world deployment.


\appendix
\section{Released Checkpoint}
\label{sec:released_checkpoint}
The publicly released PersonaPlex checkpoint\footnote{\url{https://huggingface.co/nvidia/personaplex-7b-v1}} incorporates several improvements over the experimental setup described in this paper:

\paragraph{Real Conversational Data} The released model is additionally trained on 7,303 conversations (1,217 hours) from the Fisher English corpus~\cite{cieri2004fisher} to improve natural backchanneling, expressions, and emotional responses. These conversations were annotated with prompts at varying detail levels using GPT-OSS-120B to balance generalization capability with instruction-following precision:
\begin{itemize}
    \item Minimal: ``You enjoy having a good conversation.''
    \item Topic-specific: ``You enjoy having a good conversation. Have a casual discussion about eating at home versus dining out.''
    \item Highly detailed: ``You enjoy having a good conversation. Have a reflective conversation about career changes and feeling of home. You have lived in California for 21 years and consider San Francisco your home. You work as a teacher and have traveled a lot. You dislike meetings.''
\end{itemize}

\paragraph{Synthetic Voice Generation} For data privacy, all synthetic dialogs use synthetic voices sampled from TortoiseTTS~\cite{betker2023tortoise} rather than real voice datasets described in Section~\ref{ssec:speechgen}. These synthetic voices are pitch and formant augmented using Praat~\cite{boersma2001praat} to cover a wide variety of timbres. Additionally, all synthetic dialogs (both assistant and service scenarios) use ChatterboxTTS~\cite{chatterboxtts2025} for speech generation, replacing the mixed Dia~\cite{diatts2025}/Chatterbox approach described in Section~\ref{ssec:speechgen}. Since Chatterbox provides superior speaker consistency, this unified approach yields an improved speaker similarity score of 0.65 (compared to 0.57 in Table~\ref{tab:amt_naturalness}).

\subsection{Evaluation of Released Checkpoint}
The released checkpoint demonstrates improved naturalness and conversational dynamics while maintaining the core hybrid prompting architecture and role conditioning capabilities. Compared to the experimental setup, the released checkpoint exhibits significantly increased backchannel frequency and improved pause handling, as demonstrated in Table~\ref{tab:full_duplex_bench_release}. We conducted an additional DMOS evaluation of the released checkpoint on the Full-Duplex-Bench ``User Interruption'' category using a separate annotator pool. As shown in Table~\ref{tab:released_dmos}, the released checkpoint maintains competitive naturalness relative to baseline models.

\begin{table}[h]
\centering
\caption{Released Checkpoint: Dialog Naturalness MOS (95\% C.I.) on Full-Duplex-Bench. Scores are relative within this study and not directly comparable to Table~\ref{tab:amt_naturalness}.}
\label{tab:released_dmos}
\begin{tabular}{lc}
\toprule
\textbf{Model} & \textbf{DMOS}($\uparrow$) \\
\midrule
PersonaPlex (Released) & $2.95 \pm 0.25$ \\
Gemini & $2.80 \pm 0.24$ \\
Qwen-2.5-Omni & $2.81 \pm 0.24$ \\
Freeze-Omni & $2.51 \pm 0.22$ \\
Moshi & $2.44 \pm 0.21$ \\
\bottomrule
\end{tabular}
\end{table}


\clearpage  

\bibliographystyle{IEEEbib}
\bibliography{strings,refs}
\end{document}